\documentclass{article} %
\usepackage{iclr2027_conference,times}

\usepackage{amsmath,amsfonts,bm}

\def\eqref#1{equation~\ref{#1}}

\def\1{\bm{1}}

\DeclareMathAlphabet{\mathsfit}{\encodingdefault}{\sfdefault}{m}{sl}
\SetMathAlphabet{\mathsfit}{bold}{\encodingdefault}{\sfdefault}{bx}{n}

\usepackage[colorlinks=true]{hyperref}
\definecolor{cornellred}{rgb}{0.7, 0.11, 0.11}
\definecolor{cadmiumgreen}{rgb}{0.0, 0.42, 0.24}
\definecolor{aliceblue}{rgb}{0.91, 0.94, 0.97}
\definecolor{darkblue}{rgb}{0.83, 0.89, 0.97}
\definecolor{Red7}{rgb}{0.941, 0.243, 0.243}
\definecolor{Green7}{RGB}{55, 178, 77}
\definecolor{Blue9}{rgb}{0.098,0.3,0.9}
\definecolor{oursrow}{RGB}{235,240,248}
\definecolor{tabsub}{gray}{0.35}

\hypersetup{
    colorlinks=true,
    citecolor=black,
    linkcolor=black,
    urlcolor=Blue9
}

\title{
Adaptation of Computer Use Agents\\to New Software with User Manuals }

\author{
\textbf{Chulhoon Yoon}$^{1}$,
\textbf{Kyungri Park}$^{2}$,
\textbf{Jinyeong Yun}$^{1}$, \\
\textbf{Daeyun Jeong}$^{2}$,
\textbf{Sukmin Yun}$^{2}$,
\textbf{Woohwan Jung}$^{1}$ \\
\\
$^{1}$Korea University \\
$^{2}$Hanyang University
}

\usepackage[most]{tcolorbox}
\usepackage[utf8]{inputenc} %
\usepackage[T1]{fontenc}    %
\usepackage{url}            %
\usepackage{booktabs}       %
\usepackage{amsfonts}       %
\usepackage{nicefrac}       %
\usepackage{microtype}      %
\usepackage{xcolor}         %
\usepackage{listings}
\usepackage{xspace}
\usepackage{amsmath}
\usepackage{pifont}
\usepackage{colortbl}
\usepackage{multirow}
\usepackage{makecell}
\usepackage{tabularx}
\usepackage{booktabs}
\usepackage{graphicx}
\usepackage{caption}

\iclrfinalcopy %

\newcommand{\TaskAtone}[1]{\Task^{#1}}
\newcommand{\Task}{\mathcal{I}}
\newcommand{\Plan}{\mathcal{P}}
\newcommand{\PlanAtone}[1]{\Plan^{#1}}

\newcommand{\TrajectoryLengthAt}[1]{T_{#1}}
\newcommand{\MaxTrajectoryLength}{T_{\mathrm{max}}}
\newcommand{\MinTrajectoryLength}{T_{\mathrm{min}}}

\newcommand{\Trajectory}{\tau} 
\newcommand{\TrajectoryAt}[1]{\tau_{#1}} 

\newcommand{\RewardAt}[1]{R_{#1}}
\newcommand{\SuccessReward}{R^{\mathrm{success}}} 
\newcommand{\SuccessRewardAt}[1]{R_{#1}^{\mathrm{success}}} 
\newcommand{\StepRewardAt}[1]{R_{#1}^{\mathrm{step}}} 
\newcommand{\StepWeight}{\lambda}

\newcommand{\Policy}{\pi}

\newcommand{\Eps}{\epsilon}

\newcommand{\ObsAt}[1]{o_{#1}}
\newcommand{\ActAt}[1]{a_{#1}}

\newcommand{\AdvantageAt}[2]{A_{#1,#2}}
\newcommand{\AdvantageAtone}[1]{A_{#1}}

\newcommand{\opencua}{OpenCUA\xspace}
\newcommand{\evocua}{EvoCUA\xspace}
\newcommand{\uitars}{UI-TARS-1.5\xspace}
\newcommand{\ours}{MAGE\xspace}
\newcommand{\vscode}{VSCode\xspace}

\newcommand{\Beta}{\beta}

\renewcommand{\paragraph}[1]{\noindent\textbf{#1. }}

\newcommand{\vect}[1]{\mathbf{#1}}
\newcommand{\set}[1]{\mathcal{#1}}
\begin{document}
\maketitle
\begin{abstract}
Recent advances in multimodal foundation models have significantly improved the ability to operate real-world software through graphical user interfaces (GUIs) using visual observations and natural language instructions.
However, maintaining strong performance on unseen applications remains challenging, as practical tasks often require long-horizon procedural knowledge and application-specific operation patterns.
In addition, constructing executable tasks and collecting corresponding trajectories for training still heavily rely on human-annotated data, incurring substantial cost.
In this paper, we propose \ours (Manual-augmented computer use \textbf{AG}ent \textbf{E}xecution and learning), a framework that leverages user manuals as a scalable source of procedural knowledge to improve both inference and training.
During execution, \ours retrieves task-relevant manual content and uses a planner to generate explicit procedural guidance for CUAs.
To further account for the discrepancies between manuals and actual GUI environments, such as version differences or user settings, \ours also adapts the agent policy through environment interaction.
During training, \ours converts manuals into executable task-plan pairs and uses the plans to guide environment rollouts toward task-relevant states; the resulting trajectories are then used to optimize the policy without human-authored demonstrations.
This unified design enables effective utilization of manuals for both task execution and training, reducing reliance on human-authored training data.
Experimental results across six applications demonstrate improvements in task success rates over strong baselines, highlighting the effectiveness of manual-guided planning and policy adaptation in new software environments.
We provide the code and reproduction scripts at
\url{https://github.com/ku-autoai/MAGE.git}.
\end{abstract}

\section{Introduction}
\label{sec:intro}

Recent advances in multimodal vision-language models (VLMs)~\citep{yin2024survey} have enabled Computer Use Agents (CUAs)~\citep{zhang2024large, lin2024showui, uitars, wang2025opencua} to operate software from visual observations and natural language instructions.
Unlike conventional automation methods that rely on APIs or HTML structures, these agents interact directly with graphical user interfaces (GUIs) in a human-like manner.
This enables CUAs to perform various tasks across diverse software environments.

However, existing CUAs require additional training to maintain high performance on unfamiliar applications since each application has its own interface and functionalities.
Existing adaptation~\citep{uitars, btl-ui, se-gui} typically relies on costly human annotations to define diverse executable tasks and collect corresponding action trajectories.
Although reinforcement learning~\citep{sutton1998reinforcement, silver2025welcome, shao2024deepseekmath} can reduce the need for human demonstrations, GUI tasks are inherently long-horizon: agents should sequentially uncover available options through clicks, menus, and dialogs rather than observing all relevant information at once. 
This makes it difficult to obtain sufficient successful trajectories, limiting the training signals required to improve the agent policy.

To address these limitations, we leverage user manuals as a source of procedural supervision.
Manuals describe application functionalities and the procedures required to execute them, including functions hidden behind menus, dialogs, and toolbars. 
By exposing access routes to such functions, manuals can reduce unguided exploration in long-horizon GUI tasks.
Prior approaches have also leveraged external knowledge, such as web tutorials and past interaction experiences, to guide GUI agents~\citep{rag_gui, agent_s}.
However, providing external knowledge only during execution does not directly address the challenge of collecting successful trajectories for policy adaptation.
Moreover, manuals may be abstract or differ from actual GUI environments due to software versions, user settings, or access permissions.
Thus, agents must ground procedural guidance through environment interaction and learn which actions succeed in the current GUI state.

Based on this observation, we propose \ours (Manual-augmented computer use \textbf{AG}ent \textbf{E}xecution and learning), a unified framework that leverages user manuals to improve both execution-time planning and training-time policy optimization.
During execution, MAGE retrieves task-relevant manual content and uses a planner to generate task-specific procedural plans that guide CUAs.
During training, it automatically generates diverse executable tasks and corresponding plans from user manuals.
Rather than treating generated plans as expert demonstrations, MAGE evaluates the collected trajectories and optimizes the agent policy using Group Relative Policy Optimization (GRPO).
This joint use of manuals for planning and policy learning enables application-specific adaptation without human-authored demonstrations or step-level action annotations.

Our main contributions are summarized as follows.
\begin{itemize}
    \item We propose MAGE, a unified framework that uses user manuals as procedural supervision for both execution-time planning and training-time policy learning, combining a planner with CUAs.

    \item We introduce retrieval-augmented planning, which retrieves task-relevant manual content and generates procedural plans to guide long-horizon GUI execution.
    
    \item We develop a manual-derived training method that constructs task-plan pairs and optimizes the policy from the resulting trajectories without human-authored demonstrations or step-level annotations.

    \item We evaluate MAGE across six applications and demonstrate improvements in task success rates over the corresponding baselines, examining the contributions of manual-guided planning and policy optimization.
\end{itemize}

\section{Related Work}
\label{sec:related_works}

\paragraph{Computer use agents (CUAs)}
To mitigate reliance on specific interfaces or predefined structural representations in computer environments, VLM-based agents that understand GUIs purely from visual signals have been proposed, with representative examples including \opencua~\citep{wang2025opencua} and UI-TARS~\citep{uitars}.
Training these foundation models requires large and diverse datasets.
For instance, \opencua~\citep{wang2025opencua} integrates datasets from prior works such as ShowUI~\citep{lin2024showui} and UGround~\citep{qian2025uground}, along with general supervised fine-tuning (SFT) data from the Kimi Team~\citep{team2025kimi}, and further incorporates demonstration data collected from Ubuntu, Windows, and macOS environments.
Similarly, UI-TARS~\citep{uitars} constructs additional training data by allowing human annotators to refine and correct trajectories generated by the model.
While such data-centric training strategies significantly improve model performance, adapting these models to new software environments requires additional data collection and annotation tailored to each target domain.
This process incurs substantial cost, thereby limiting the scalability and practical applicability of CUAs.

\paragraph{Reducing human supervision for CUA adaptation}
Prior work has explored reducing human supervision for CUA adaptation by augmenting or reusing existing datasets for training, or by leveraging external knowledge to guide task execution.
BTL-UI~\citep{btl-ui} augments existing datasets with ROI annotations,
while ZeroGUI~\citep{yang2025zerogui} and DART~\citep{dart} leverage benchmark-derived tasks for reinforcement learning.
Meanwhile, RAG-GUI~\citep{rag_gui} trains a VLM to generate guidelines from web tutorials, guiding a frozen CUA at inference time, and Agent S~\citep{agent_s} uses external knowledge for hierarchical planning.
However, data-driven approaches still rely on task-specific data, while knowledge-based approaches primarily assist execution at inference time without adapting the CUA's policy.
In contrast, \ours leverages user manuals for both planning and policy optimization, improving task execution without repeatedly generating plans throughout an episode.

\section{Method}
\ours is a manual-augmented adaptation framework for CUAs. 
Given a target program and its user manual, MAGE uses the manual as procedural supervision in two complementary ways: it retrieves task-relevant instructions to guide long-horizon execution, and it generates executable task-plan pairs to support policy training. 
These two components address the main bottlenecks in GUI adaptation: reliable task execution and scalable collection of successful trajectories.

\subsection{Problem Definition}
Given a natural language task instruction $\Task$, our goal is to complete the task $\Task$ by performing a sequence of actions. 
At each time step $t$, the agent observes a screenshot  $\ObsAt{t}$, generates an action $\ActAt{t}$, and receives the next observation $\ObsAt{t+1}$.
This process forms a trajectory $\Trajectory = \{(\ObsAt{1}, \ActAt{1}), (\ObsAt{2}, \ActAt{2}), \dots, (\ObsAt{T}, \ActAt{T})\}$ where $T$ is the length of the trajectory.
The action space includes common mouse and keyboard actions such as click, right click, double click, scroll, drag, and typing.
Please refer to Appendix~\ref{sec:actionspace} for the detailed descriptions of each action and its arguments.

\subsection{Task Execution with Retrieval-Augmented Planning}
\label{sec:task_execution}
\begin{figure}[h]
    \vspace{-0.15in}
    \centering
    \includegraphics[width=1\linewidth]{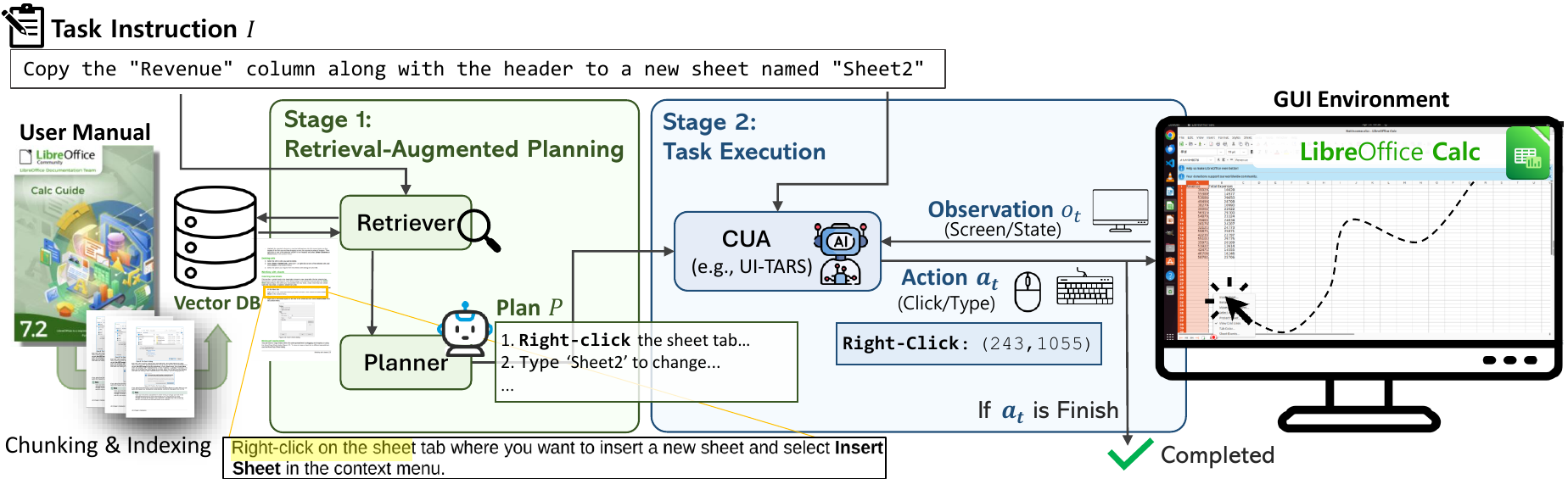}
    \caption{Task execution with user manuals. MAGE retrieves relevant manual sections, then generates a retrieval-augmented plan, and finally performs GUI task execution conditioned on the plan.}
    \label{fig:overview_execution}
    \vspace{-0.05in}
\end{figure}

CUAs execute a task by repeatedly selecting the most appropriate action $\ActAt{t}$ based on the current observation $\ObsAt{t}$ and the past trajectory $\Trajectory_{<t} = \{(\ObsAt{j},\ActAt{j})|1 \leq j <t\}$.
However, in many GUI environments, the information required for the task completion is not fully visible in the current screenshot.
Important functions are often hidden behind menus, dialogs, toolbars, or keyboard shortcuts, degrading the performance of agents for long-horizon tasks. 
To address the problem, we propose retrieval-augmented planning which uses procedural guidance from user manuals during task execution.

As shown in Figure~\ref{fig:overview_execution}, our task execution consists of two stages.
In the first stage, \ours performs retrieval-augmented planning, following the retrieval-augmented generation framework~\citep{lewis2020retrieval}.
It first retrieves relevant manual sections to provide detailed procedural information to the planner and generates an execution plan consisting of a sequence of actions with specific information such as button names and keyboard shortcuts.
In the next stage, \ours executes the task similarly to conventional CUAs but with the execution plan.

\paragraph{Retrieval-augmented planning}
\ours retrieves manual sections relevant to $\Task$ and generates a task-specific plan.
Let $\set{M} = \{m_1, \ldots, m_{|M|}\}$ denote the set of manual sections.
We encode each manual section $m_i$ using an embedding model as $\vect{h}_i=E(m_i)$.
The embedding vectors are indexed in a vector database before executing the CUA.
To retrieve the relevant section, \ours encodes the task instruction as $\vect{q}=E(\Task)$ and finds the top-$K$ sections $\set{M}_\Task$ with the highest cosine similarity $\cos(\vect{q}, \vect{h}_i)$. 
The planner then takes the task instruction $\Task$, the relevant sections $\set{M}_\Task$, and the initial observation $\ObsAt{1}$ as inputs, and generates a plan $\Plan$ that includes step-by-step procedures with specific actions.
We use a multimodal LLM as the planner and provide its prompt in Appendix~\ref{sec:plannerprompt}.

\paragraph{Task execution with the generated plan}
Given the generated plan, \ours executes the task by conditioning the CUA on the task instruction $\Task$, the plan $\Plan$, the current observation $\ObsAt{t}$, and the past trajectory $\Trajectory_{<t}$.
At each time $t$, the agent selects the next action as 
\begin{equation}
    \ActAt{t} \sim  \pi_{\theta}(\cdot|\ObsAt{t},\Trajectory_{<t}, \Plan, \Task).
\end{equation}
The plan remains fixed throughout the episode, while the CUA selects actions based on the current observation and interaction history.

\subsection{Training CUAs with Manual-Derived Tasks}
\begin{figure}[h]
    \vspace{-0.15in}
    \centering
    \includegraphics[width=1\linewidth]{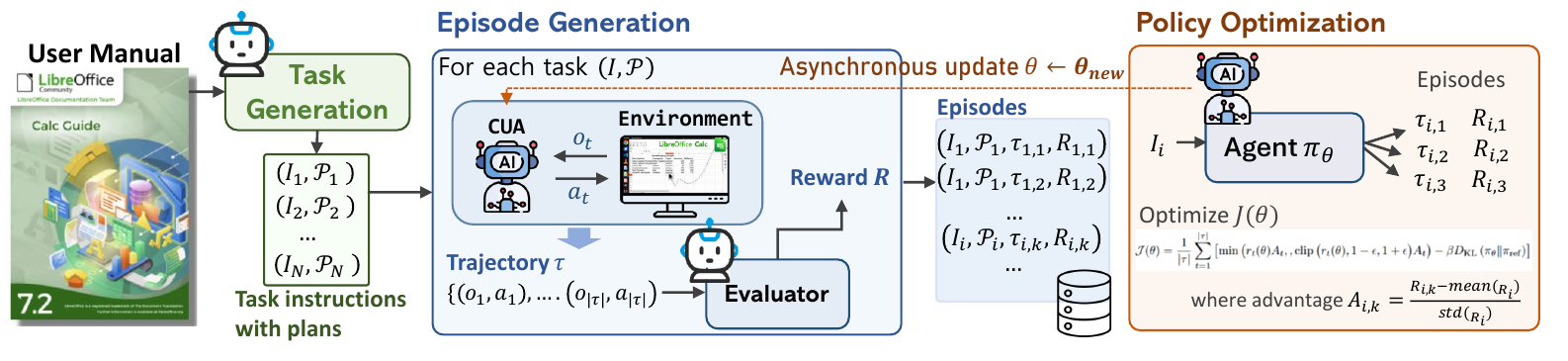}
    \caption{Training CUAs with a user manual. MAGE uses manual-derived task-plan pairs to collect episodes for policy optimization.}
    \label{fig:overview_training}
    \vspace{-0.05in}
\end{figure}

Retrieval-augmented planning provides procedural guidance, but manuals do not fully specify the actual GUI state of a target environment. 
In addition, interface elements can be hidden, customized, or restricted depending on application settings, workspace configurations, and user permissions. 
Thus, even when a manual describes a valid procedure, the agent should learn how to execute it robustly in the observed environment.
To address this problem, we utilize user manuals not only for execution-time planning but also for policy training.

As shown in Figure~\ref{fig:overview_training}, the training pipeline consists of three stages: (1) task generation, (2) episode generation, and (3) policy optimization. 
In the first stage, MAGE uses the manual to construct a set of executable task-plan pairs. 
This stage is performed once before training begins.
Second, the current policy executes these task-plan pairs in the target GUI environment to collect episodes. 
Third, the collected episodes are evaluated and used to optimize the policy via GRPO~\citep{shao2024deepseekmath}. 
The second and third stages are repeated iteratively, so that each updated policy produces new episodes for subsequent optimization.

\subsubsection{Task Generation}
In this stage, \ours generates a wide range of executable task instructions from the manual.
Since policy optimization requires successful trajectories, training becomes ineffective when the initial policy rarely completes long-horizon GUI tasks.
To address this, we generate plans together with tasks to guide execution and steer rollouts toward task-completing trajectories.

For each manual section, \ours uses both the textual content and images as inputs.
A multimodal LLM interprets the functional descriptions and examples in the section, and generates a setup task $\TaskAtone{\mathrm{setup}}$, a setup plan $\PlanAtone{\mathrm{setup}}$, a main task $\TaskAtone{\mathrm{main}}$, and a main plan $\PlanAtone{\mathrm{main}}$.
The setup task is an optional preliminary task that configures the initial GUI state required to execute the main task, while the main task is the final task that requires using the functionality described in the corresponding manual section.
Each plan provides step-by-step procedural guidance for executing its corresponding task.

By explicitly separating setup tasks, the framework can generate and execute diverse tasks without requiring manually prepared initial environments.
The resulting setup and main tasks naturally involve common operations in the target program, such as entering values into spreadsheet cells or navigating application menus.
Training on these tasks exposes the agent to frequently used program operations while covering diverse functionalities and usage procedures described in the manual.
The prompt template for task generation is provided in Appendix~\ref{app:prompt_task_generation}.

\subsubsection{Episode Generation}
\label{sec:episode_generation}

In this stage, trajectories are constructed by executing the generated task pairs in the GUI environment. 
The trajectory collection follows the interaction process described in Section~\ref{sec:task_execution}. 
When a setup task is included in a task pair, the agent first executes the setup task to configure the required GUI state, and proceeds to the main task only upon successful completion. 
If there is no setup task, the agent directly executes the main task from the initial state. 
Each task execution generates a trajectory $\Trajectory = \{(\ObsAt{1}, \ActAt{1}), (\ObsAt{2}, \ActAt{2}), \dots, (\ObsAt{T}, \ActAt{T})\}$ consisting of a sequence of observations and actions.

To determine task success, we adopt the LLM-as-a-judge paradigm and employ a vision-language model (VLM) as an evaluator. 
The evaluator assigns a binary success reward $\SuccessReward \in \{0,1\}$ to each trajectory, where $\SuccessReward = 1$ if the task is judged as successful and $\SuccessReward = 0$ otherwise. 
However, prior work~\citep{yang2025zerogui} has shown that evaluating full trajectories can be unreliable, as important details may be overlooked and hallucinations may lead to incorrect judgments. 
To mitigate this issue, we determine task success from the task instruction and the last five screenshots of the trajectory.
This design aims to reduce susceptibility to hallucinations and evaluation cost.
We provide a detailed analysis of the evaluator in Appendix~\ref{appendix:evaluator_validation}.

For policy optimization, we construct training episodes from the collected task executions. Setup and main tasks are handled as separate training instances rather than being merged into a single trajectory.
Accordingly, each task execution produces a separate training episode, defined as a tuple $(\Task, \Plan, \Trajectory, \SuccessReward)$. Each task is executed multiple times to collect a diverse set of episodes, which are subsequently used for GRPO-based policy optimization.

\subsubsection{Policy Optimization with Manual-Derived Episodes}
\label{subsec:policy_opt}
In our adaptation setting, the amount of interaction data and available computation are  limited compared to general foundation model training.
Thus, we utilize GRPO (Group Relative Policy Optimization)~\citep{shao2024deepseekmath} to optimize the CUA from collected episodes instead of using other reinforcement learning algorithms requiring a separate value model, such as PPO~\citep{DBLP:journals/corr/SchulmanWDRK17}.

\paragraph{Round-based training}
GRPO requires multiple rollouts per task to estimate group-level statistics, and learning becomes more effective as the proportion of episodes with positive rewards increases. 
However, generating new rollouts at every training step incurs substantial computational cost.
We define a \textit{round} as a cycle of episode generation and GRPO-based training.
To address this, we propose a round-based training strategy that builds on an offline learning setup by reusing existing data while incrementally incorporating rollouts generated by the latest policy. 
In each round, newly generated episodes are added to the existing dataset and used for subsequent training. 
By repeating this process, rollouts with higher success rates are progressively collected, enabling the dataset to expand and the policy to improve over time.

\paragraph{Episode sampling}
Instead of computing expectations over all episodes within a group, we adopt a sampling-based approximation. 
For each task, one episode is randomly sampled for training, while its advantage is computed using all available episodes in the group.
In this process, setup tasks and main tasks are treated as distinct, independent task instances, and sampling is performed separately for each task.
Due to the dependency between setup and main tasks, not all attempts result in valid episodes, and multiple trials for a given task may yield only a limited number of episodes. 
In particular, when all episodes for a given task have identical rewards or when only a single episode is available, the reward variance becomes zero, making advantage estimation ill-defined.
To address this issue, we exclude episode groups with zero reward variance from policy optimization.
This approach reduces computational overhead by reusing previously collected episodes rather than generating new rollouts at every training step.

\paragraph{Reward design}
We define the reward function based on both task success and execution efficiency. 
Let $\TrajectoryAt{i}$ denote the $i$-th trajectory with length $\TrajectoryLengthAt{i}$, and $\RewardAt{i}$ denote the corresponding trajectory-level reward.
We use the success reward $\SuccessRewardAt{i}$ defined in Section~\ref{sec:episode_generation}.
To encourage efficient execution, we define a step-based reward that is only applied upon success:
\begin{equation}
\StepRewardAt{i} =
\begin{cases}
\frac{
\MaxTrajectoryLength - \TrajectoryLengthAt{i}
}{
\MaxTrajectoryLength - \MinTrajectoryLength
}, 
& \text{if } \SuccessRewardAt{i} = 1, \\
0, & \text{otherwise}.
\end{cases}
\end{equation}
Here, $\MaxTrajectoryLength$ and $\MinTrajectoryLength$ denote the maximum and minimum trajectory lengths among episodes of the same task. 
The step-efficiency reward is set to zero when $\MaxTrajectoryLength = \MinTrajectoryLength$.
The total reward is given by:
\begin{equation}
\RewardAt{i}
=
\SuccessRewardAt{i}
+
\StepWeight \StepRewardAt{i},
\end{equation}
where $\StepWeight$ is a weighting coefficient.

\paragraph{Advantage estimation}
Advantage is computed using the set of episodes corresponding to the same task. 
Let $\mathcal{D}_u = \{\tau_i\}_{i=1}^N$ denote the set of trajectories for task $u$. 
We compute the normalized advantage as:
\begin{equation}
\AdvantageAt{i}{k}
=
\frac{
\RewardAt{i,k} - \operatorname{mean}(\mathcal{R}_i)
}{
\operatorname{std}(\mathcal{R}_i)
}.
\label{eq:Advantage}
\end{equation}
Here, $i$ indexes the task, and $k$ indexes the $k$-th trajectory (episode) associated with task $i$. 
$\mathcal{R}_i = \{\RewardAt{i,k}\}_{k=1}^{K_i}$ denotes the set of rewards for all trajectories corresponding to task $i$, and $\operatorname{mean}(\mathcal{R}_i)$ and $\operatorname{std}(\mathcal{R}_i)$ represent the mean and standard deviation over this set, respectively.

\paragraph{Objective function}
We optimize the policy using a clipped GRPO objective. 
Let $\Policy_{\theta}$ denote the current policy and $\Policy_{\theta_{\mathrm{old}}}$ the old policy.
At the beginning of each training epoch, we reset the old policy to the current policy.
For a sampled trajectory ($\tau$), we define the probability ratio at step $t$ as
\begin{equation}
r_t(\theta)
=
\frac{
\Policy_{\theta}
\left(
\ActAt{t} \mid \Task, \Plan, \Trajectory_{<t}, \ObsAt{t}
\right)
}{
\Policy_{\theta_{\mathrm{old}}}
\left(
\ActAt{t} \mid \Task, \Plan, \Trajectory_{<t}, \ObsAt{t}
\right)
}.
\end{equation}
The objective for a sampled episode is:
\begin{equation}
\label{eq:objective}
\mathcal{J}(\theta)
=
\frac{1}{|\Trajectory|}
\sum_{t=1}^{|\Trajectory|}
\left[
\min \left(
r_t(\theta)\AdvantageAtone{t},
\operatorname{clip}\big(r_t(\theta), 1-\Eps, 1+\Eps\big)\AdvantageAtone{t} \right)
-
\Beta D_{\mathrm{KL}}\left(\Policy_{\theta}\|\Policy_{\mathrm{ref}}\right)
\right],
\end{equation}
where $\epsilon$ is the clipping parameter, $\beta$ controls the KL penalty, and $\Policy_{\mathrm{ref}}$ is the initial CUA policy before adaptation, kept frozen throughout training.

\section{Experiments}
\label{sec:experiments}

\subsection{Experimental Setup}

\noindent{\bf Benchmarks and applications. }
We evaluate \ours on the OSWorld benchmark~\citep{osworld} for five applications: LibreOffice Calc, Impress, and Writer, GIMP, and \vscode.
These applications require diverse multi-step GUI interactions including spreadsheet editing, slide modification, document editing, image editing, and code-related operations.
Since OSWorld is a widely used CUA benchmark, tasks on these applications may already be in CUA training data. 
To examine whether MAGE can also adapt CUAs to applications with potentially limited coverage in their training, we construct 30 new evaluation tasks for Slack, a workplace communication platform.

\paragraph{Models}
We evaluate our framework with two CUA backbones: \uitars (7B)~\citep{uitars} and \evocua (8B)~\citep{xue2026evocua}.
For retrieval-augmented planning, we use Qwen3-Embedding-0.6B~\citep{zhang2025qwen3} for manual retrieval and Qwen3.6 (27B) for plan generation.
At the beginning of each task, 
the planner generates a task-specific plan conditioned on the task instruction, initial screenshot, and retrieved manual content.
The generated plan remains fixed throughout execution.
We use GPT-5.4 for manual-based task generation and trajectory evaluation during training.

\paragraph{Training and planning setup}
For LibreOffice applications, we use the official LibreOffice user manuals\footnote{\url{https://documentation.libreoffice.org/en/english-documentation/}} for both retrieval-augmented planning and manual-derived task construction.
For GIMP\footnote{\url{https://docs.gimp.org/3.2/en/}} and \vscode\footnote{\url{https://code.visualstudio.com/docs}}, we use their official documentation, and for Slack, we use the Slack Help Center\footnote{\url{https://slack.com/help}}.
We use the Round-based Training strategy described in Section~\ref{subsec:policy_opt} for policy optimization.
Training is performed separately for each application to evaluate application-specific adaptation.
Within each application, the training configuration differs between the \uitars and \evocua backbones.
Detailed backbone-specific settings, including the number of rounds and episodes generated per round, are provided in Appendix~\ref{appendix:training_details}.

\noindent{\bf Evaluation setup. }
To account for the inherent stochasticity of CUAs, we evaluate our main models over five independent runs on both OSWorld and Slack and report the mean and standard deviation.

For OSWorld, we evaluate performance across five applications. Because the number of tasks differs across applications, the overall Success Rate (SR) is micro-averaged by dividing the total number of successful tasks by the total number of tasks across all applications.
For the backbone models used in MAGE, the five runs include results from the official OSWorld dashboard (two for UI-TARS-1.5 and one for EvoCUA).

To evaluate \ours on Slack, we construct a separate test set consisting of 30 task pairs across 15 themes.
For Slack evaluation, we use Gemini 3.1 Pro Preview to determine task success from the task instruction and the last five screenshots of each trajectory.
When a setup task is present, the agent must complete it before attempting the main task. 
We report the Success Rate (SR) across all 30 task pairs based on successful completion of the main task.
Further details on the Slack evaluation are provided in Appendix~\ref{appendix:slack_eval}.

\begin{table}[t]
\caption{
Task success rates (\%) on OSWorld applications under a 50-step budget.
For models evaluated over multiple runs, we report mean $\pm$ standard deviation.
Bold and underlined values indicate the highest and second-highest mean success rates, respectively, among systems with base CUAs of 8B parameters or fewer.
}
\label{tab:main_results_osworld}
\centering
\small
\setlength{\tabcolsep}{3pt}

\newcommand{\msd}[2]{#1\,{\scriptsize $\pm$\,#2}}
\newcommand{\bmsd}[2]{\textbf{#1}\,{\scriptsize $\pm$\,#2}}
\newcommand{\umsd}[2]{\underline{#1}\,{\scriptsize $\pm$\,#2}}

\resizebox{\linewidth}{!}{%
\begin{tabular}{@{}l|cc|ccccc|c@{}}
\toprule
\multicolumn{1}{c|}{\textbf{Model}} &
\textbf{Size} &
\textbf{Runs} &
\multicolumn{1}{c}{\textbf{Calc}} &
\multicolumn{1}{c}{\textbf{Impress}} &
\multicolumn{1}{c}{\textbf{Writer}} &
\multicolumn{1}{c}{\textbf{GIMP}} &
\multicolumn{1}{c|}{\textbf{\vscode}} &
\multicolumn{1}{c}{\textbf{Overall}} \\
\hline

o3
& - & 1
& 8.51
& 4.26
& 21.74
& 38.46
& 21.74
& 15.66 \\

Claude-Sonnet-4.5
& - & 1
& 65.96
& 57.49
& 65.22
& 57.69
& 69.57
& 62.66 \\

\hline

UI-TARS-DPO~\citep{uitars}
& 72B & 1
& 12.77
& 25.45
& 43.48
& 61.54
& 47.83
& 33.11 \\

OpenCUA~\citep{wang2025opencua}
& 32B & 3
& \msd{13.48}{1.23}
& \msd{38.28}{4.26}
& \msd{40.58}{2.51}
& \msd{69.23}{3.85}
& \msd{55.07}{10.94}
& \msd{38.75}{1.52} \\

OpenCUA~\citep{wang2025opencua}
& 72B & 1
& 31.91
& 44.66
& 60.87
& 76.92
& 73.91
& 52.40 \\

GUI-Owl-1.5~\citep{ye2025mobile}
& 32B & 1
& 48.94
& 63.81
& 60.87
& 76.92
& 86.96
& 64.45 \\

EvoCUA~\citep{xue2026evocua}
& 32B & 1
& 55.32
& 59.49
& 69.57
& 76.92
& 86.96
& 66.24 \\

\hline

OpenCUA~\citep{wang2025opencua}
& 7B & 3
& \msd{13.12}{2.15}
& \msd{32.60}{4.33}
& \msd{33.33}{2.51}
& \msd{43.59}{5.88}
& \msd{46.38}{2.51}
& \msd{30.82}{0.27} \\

TianXi-Action~\citep{cheng2025evolving}
& 7B & 2
& \msd{6.38}{0.00}
& \msd{38.24}{0.05}
& \bmsd{54.35}{3.07}
& \msd{55.77}{2.72}
& \umsd{67.39}{3.07}
& \msd{38.24}{1.29} \\

\uitars~\citep{uitars}
& 7B & 5
& \msd{14.47}{6.45}
& \msd{38.15}{3.38}
& \msd{40.87}{2.38}
& \msd{53.85}{3.85}
& \msd{56.52}{5.32}
& \msd{36.83}{3.09} \\

\evocua~\citep{xue2026evocua}
& 8B & 5
& \msd{32.77}{2.43}
& \msd{35.98}{3.38}
& \msd{46.09}{5.83}
& \umsd{80.00}{3.22}
& \msd{66.09}{1.94}
& \umsd{47.54}{0.96} \\

\hline

\rowcolor{blue!15}
\textbf{MAGE} (\uitars)
& 7B & 5
& \umsd{34.04}{3.69}
& \bmsd{45.39}{1.90}
& \msd{44.35}{1.94}
& \msd{60.77}{5.70}
& \msd{66.09}{3.64}
& \msd{47.31}{2.02} \\

\rowcolor{blue!15}
\textbf{MAGE} (\evocua)
& 8B & 5
& \bmsd{37.02}{1.90}
& \umsd{45.32}{3.56}
& \umsd{50.43}{4.96}
& \bmsd{81.54}{1.72}
& \bmsd{76.52}{2.38}
& \bmsd{53.67}{1.73} \\

\bottomrule
\end{tabular}%
}
\end{table}

\subsection{Main Results}
\noindent{\bf OSWorld. }
As shown in Table~\ref{tab:main_results_osworld}, MAGE consistently improves performance across all five applications and both backbone models.
To assess the statistical significance of the overall success rate improvement, we conduct a two-sided Welch's t-test.
The improvements are statistically significant for both \uitars ($p = 4.13 \times 10^{-4}$) and \evocua ($p = 3.63 \times 10^{-4}$).
Notable improvements are observed in Calc and \vscode, with gains of 19.57 \%p for \uitars on Calc and 10.43\%p for \evocua on \vscode.
Many tasks in Calc and \vscode can be performed using keyboard shortcuts or application-specific commands, potentially reducing the need for complex menu navigation.
We hypothesize that MAGE benefits from retrieving these procedures from user manuals and incorporating them into its plans, helping agents execute tasks more effectively.

\paragraph{Slack}
\begin{figure}[t!]
\centering

\begin{minipage}[t]{0.46\textwidth}
\vspace{0pt}
\captionof{table}{
Mean SR comparison on the Slack.
}
\label{tab:slack}
\centering
\small

\begin{tabular}{lc}
\toprule
\textbf{Model} & \textbf{Mean SR $\pm$ SD (\%)} \\
\midrule
UI-TARS-1.5
& $26.67 \pm 5.27\%$ \\
EvoCUA
& $32.67 \pm 4.94\%$ \\
\rowcolor{blue!15}
\ours (UI-TARS-1.5)
& $\mathbf{49.33 \pm 11.88\%}$ \\
\rowcolor{blue!15}
\ours (EvoCUA)
& $\mathbf{51.33 \pm 8.03\%}$ \\
\bottomrule
\end{tabular}
\end{minipage}
\hfill
\begin{minipage}[t]{0.50\textwidth}
\vspace{0pt}
\captionof{table}{
Ablation study on OSWorld Calc.
}
\label{tab:ablation}
\centering
\small

\begin{tabular}{cc|cc}
\toprule
\textbf{RAP} & \textbf{PO} & \textbf{SR (\%)} & \textbf{Improvement} \\
\midrule
- & - & 14.47 & - \\
- & \checkmark & 14.89 & +0.42 \\
\checkmark & - & 25.53 & +11.06 \\
\rowcolor{blue!15}
\checkmark & \checkmark & \textbf{34.04} & \textbf{+19.57} \\
\bottomrule
\end{tabular}
\end{minipage}

\end{figure}
As shown in Table~\ref{tab:slack}, MAGE improves the success rate on Slack by 22.66\%p and 18.66\%p for \uitars and \evocua, respectively.
Both improvements are statistically significant under a two-sided Welch's t-test over five independent runs ($p = 0.0094$ and $p = 0.0035$, respectively).
Notably, these gains are larger than those observed on OSWorld for both backbones.
The relatively low performance of both backbone models on Slack may reflect limited exposure to Slack-specific tasks during training. Although their exact training data coverage is unknown, Slack is absent from several widely used CUA benchmarks.
By leveraging user manuals for planning and policy optimization, MAGE may compensate for this lack of application-specific training experience, leading to larger performance gains on Slack.
These results suggest its potential for adapting to new applications with limited prior training data.

\subsection{Analysis}
We provide additional analyses of \ours using \uitars as the backbone.

\subsubsection{Ablation Study}
We conduct an ablation study to evaluate the contributions of \ours's two key components: Retrieval-augmented Planning (RAP) during execution and Policy Optimization (PO) during training.
The baseline and \ours results are averaged over five independent runs, while the other variants are evaluated using a single run.
With PO alone, the success rate is 0.42 percentage points (pp) higher than the baseline, whereas with RAP alone, it is 11.06 pp higher than the baseline.
Combining RAP and PO provides an additional gain of 8.51 pp over RAP alone, resulting in a total improvement of 19.57 pp over the baseline.
These results highlight the effectiveness of manual-guided planning and suggest that PO provides additional benefits when combined with RAP.

\begin{figure}[h]
    \vspace{-0.15in}
    \centering
    \includegraphics[width=0.97\linewidth]{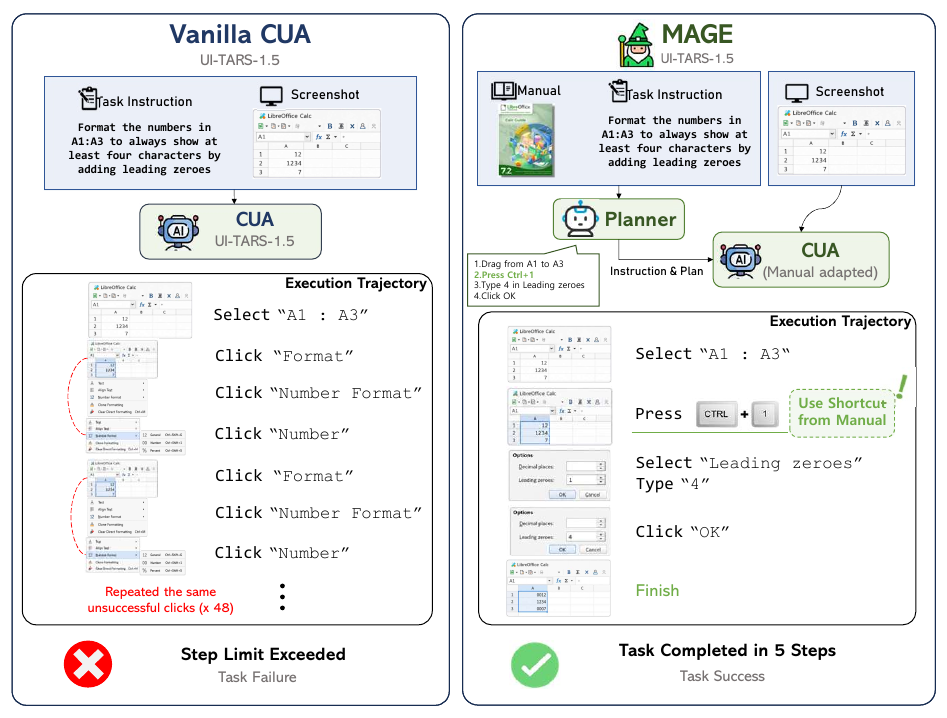}
    \caption{Qualitative comparison of GUI execution on the same task before and after applying MAGE.}
    \label{fig:shortcut}
    \vspace{-0.05in}
\end{figure}

\subsubsection{Qualitative Evaluation}

As shown in figure~\ref{fig:shortcut}, the vanilla CUA repeatedly executes a sequence of three distinct clicks 16 times, resulting in 48 unnecessary actions before reaching the maximum step budget of 50 and failing the task. 
In contrast, \ours retrieves five relevant manual sections (Numbers as Text, Date and Time, Leading Zeroes, Manual Line Break, and Introduction) and generates a task execution plan based on the retrieved information.
The resulting plan instructs the agent to use the \texttt{Ctrl+1} shortcut to open the Format Cells dialog, enabling it to successfully complete the task in only five steps.

Additionally, when the planner generates a plan without access to the retrieved manual sections, the agent still completes the task in seven steps by opening the Format Cells dialog through mouse clicks.
These results show that providing a plan even without user manuals helps the agent avoid repetitive actions.
Incorporating user manuals further enables the agent to leverage documented application-specific functionalities.
This allows the agent to execute tasks more efficiently.

\subsubsection{Cost Analysis}
\begin{table}[!t]
\caption{Adaptation cost of \ours across applications using \uitars as the base agent.}
\label{tab:adaptation_cost}
\centering
\small
\setlength{\tabcolsep}{4pt}

\resizebox{0.75\linewidth}{!}{%
\begin{tabular}{l|rrrrr|r}
\toprule
& \textbf{Calc} & \textbf{Impress} & \textbf{Writer} & \textbf{GIMP} & \textbf{Slack} & \textbf{Average} \\
\midrule
API Cost (\$)
& 79.80 & 30.66 & 33.74 & 28.95 & 15.99 & 37.83 \\

Episode Generation (h)
& 70.90 & 26.63 & 26.13 & 26.63 & 33.35 & 36.73 \\

Policy Optimization (h)
& 24.24 & 8.44 & 8.73 & 4.24 & 3.80 & 9.89 \\
\bottomrule
\end{tabular}%
}
\end{table}

To quantify the resources required to adapt \ours to a new application, we analyze the adaptation cost when using UI-TARS-1.5 as the base agent, as summarized in Table~\ref{tab:adaptation_cost}. 
Specifically, we measure the monetary API cost, the wall-clock times required for episode generation, and policy optimization.
On average across the five applications reported in the table, adapting \ours to a new application requires 46.62 hours on three NVIDIA H100 GPUs and approximately \$37.83 in API costs.

To contextualize the cost of training data generation, OpenCUA~\citep{wang2025opencua} employed 634 annotators
to collect approximately 22,000 tasks over six months.
AgentTrek~\citep{xu2025agenttrek} generated approximately 10,400 verified trajectories at an estimated API cost of \$5,726.7. 
Although these figures represent different data collection scales, MAGE demonstrates the feasibility of adapting CUAs to individual applications at a relatively low API cost without human-authored demonstrations.

\section{Conclusion}
\label{sec:conclusion}
We presented \ours, a framework that leverages user manuals for both task execution with manual-guided planning and training with manual-derived tasks.
In the execution pipeline, \ours retrieves task-relevant manual sections and uses a planner to generate a plan conditioned on the task instruction and the initial screen state.
The generated plan is provided to CUAs, together with the task instruction and trajectory history, allowing them to incorporate external procedural knowledge during action generation.
In the training pipeline, \ours constructs manual-derived tasks from function-level manual sections, collects self-generated trajectories in the GUI environment, assigns rewards through an automatic evaluator, and performs GRPO-based policy optimization. 
This design integrates manual-based task construction, agent--environment interaction, reward assignment, and policy optimization without requiring human-authored expert demonstrations.

Our experiments demonstrate that incorporating user manuals into planning and policy optimization improves task success across multiple applications. 
These results suggest that user manuals can be used not only as human-facing documentation, but also as scalable supervision for building more capable CUAs.

\newpage

\subsection*{AI Use Statement}
In this work, we used generative AI tools to generate synthetic datasets; design and provide feedback on research methodology and experiments; implement methods; assist with translation; clean and reformat datasets; and support qualitative and thematic data analysis. 
We did not use generative AI tools to propose or refine hypotheses, develop theoretical models or conceptual frameworks, formulate mathematical claims, provide critical ingredients for proving mathematical claims, or interpret experimental results.

All AI-assisted outputs were reviewed and validated by the authors. 
In particular, AI-generated tasks and plans were manually examined for usefulness, feasibility, requirement coverage, procedural correctness, executability, and scope safety. 
AI-assisted implementations were tested through controlled experiments and benchmark evaluations, and automatically generated trajectory assessments were cross-checked against the official rule-based evaluators. 
The authors independently analyzed and interpreted the experimental results. 
We take full responsibility for the final content of this work, including all text, claims, code, data, and other artifacts produced with the aid of generative AI.

\subsection*{Ethics Statement}
This work aims to reduce the human demonstration effort required to adapt computer-using agents to target applications. Such agents may also cause unintended operations, data loss, or unauthorized actions. Our experiments evaluate task execution in designated test environments; the results do not establish safe deployment in unrestricted settings. Practical deployment should incorporate restricted permissions, user confirmation for sensitive actions, and appropriate logging. The safeguards discussed here are deployment recommendations rather than guarantees provided by our method.

\subsection*{Reproducibility Statement}
The main text and appendix describe the task-generation procedure, retrieval-augmented planning, policy optimization, and evaluation protocols. The appendix provides prompt templates, training configurations, action spaces, and additional analyses. To support reproducibility, we release the training and execution code, reproduction scripts, tasks, trajectories, and trained model checkpoints through the repository linked in the abstract. Because Slack requires account authentication, its execution trajectories may contain account-identifying information; we will therefore release them after removing such information.

\bibliography{iclr2027_conference}
\bibliographystyle{iclr2027_conference}
\appendix
\newpage

\lstset{
  basicstyle=\ttfamily\small,
  breaklines=true,
  columns=fullflexible,
  keepspaces=true,
  showstringspaces=false
}

\section{Experimental Details}

\subsection{Implementation Details}

\begin{table*}[t]
\centering
\caption{Data generation and execution statistics of MAGE across six applications.}
\label{tab:task_analysis}
\small
\setlength{\tabcolsep}{5pt}

\resizebox{\textwidth}{!}{%
\begin{tabular}{l|r|rr|r|rrr}
\toprule
& \multicolumn{1}{c|}{\textbf{Manual}} &
\multicolumn{2}{c|}{\textbf{Tasks}} &
\multicolumn{1}{c|}{\textbf{Plan}} &
\multicolumn{3}{c}{\textbf{Trajectory}} \\
\cmidrule(lr){2-2}
\cmidrule(lr){3-4}
\cmidrule(lr){5-5}
\cmidrule(lr){6-8}
\textbf{Application} &
\multicolumn{1}{c|}{\textbf{\# of Sections}} &
\textbf{w/o Setup} &
\textbf{All} &
\multicolumn{1}{c|}{\textbf{Avg. Length}} &
\textbf{Trajectories} &
\textbf{Avg. Steps} &
\textbf{SR (\%)} \\
\midrule
Calc    & 536   & 34  & 228 & 7.47 & 1,665 & 29.74 & 52.13 \\
Impress & 107   & 66  & 102 & 5.53 & 609   & 30.71 & 47.62 \\
Writer  & 537   & 33  & 105 & 6.25 & 710   & 28.45 & 46.76 \\
GIMP    & 1,870 & 21  & 100 & 8.31 & 536   & 35.72 & 35.26 \\
\vscode & 1,636 & 121 & 260 & 6.78 & 1,417 & 40.52 & 21.88 \\
Slack   & 2,045 & 34  & 60  & 6.32 & 339   & 35.26 & 31.27 \\
\midrule
\textbf{Average} &
\textbf{1,121.83} &
\textbf{51.50} &
\textbf{142.50} &
\textbf{6.78} &
\textbf{879.33} &
\textbf{33.40} &
\textbf{39.15} \\
\bottomrule
\end{tabular}%
}
\end{table*}
Table~\ref{tab:task_analysis} summarizes the statistics of manual sections, generated tasks, plans, and trajectories for each application.

\paragraph{User manuals}
We collect publicly available English user manuals for six applications: LibreOffice Calc, Impress, Writer, GIMP, \vscode, and Slack.
For LibreOffice, we use Calc Guide 7.1 (May 2021), Impress Guide 7.6 (October 2023), and Writer Guide 7.3 (March 2022), respectively.
For GIMP, we use the official GIMP 3.2 documentation. 
For \vscode and Slack, we collect documentation from their respective official websites, including the \vscode documentation and Slack Help Center.

\paragraph{Manual section segmentation}
We segment user manuals into section-level chunks based on their document structure.
For LibreOffice, we divide the PDF manuals according to their tables of contents. For web-based documentation, including GIMP, \vscode, and Slack, we use HTML headings to define section boundaries. 

\paragraph{Task generation}
Before task generation, we exclude manual sections covering introductory material, application overviews, version updates, file sharing, and operations requiring external resources or integration with other applications.
For each retained section, we generate up to two tasks. 
We then sample a subset of the generated tasks to keep the number of training tasks manageable.
For Slack, we further exclude features unavailable in a free workspace with administrator privileges.

\paragraph{Trajectory generation}
For each generated task, we collect trajectories over five independent runs.
We report the number of generated trajectories, their average number of steps, and the training success rate (SR).

\paragraph{Agent inference setting}
Although the formulation conditions action generation on the full trajectory, our implementation uses only the five most recent screenshots as trajectory context.
For fair comparison, we retain the OSWorld-provided configurations for \uitars and \evocua during both training and inference.
For both trajectory generation and evaluation, we use a maximum interaction budget of 50 steps.

\subsection{Training Details}
\label{appendix:training_details}

Unless otherwise specified, training experiments are conducted on a server equipped with three NVIDIA H100 GPUs.

\paragraph{Common hyperparameters}
For both backbones, we set the step-efficiency reward weight to $\lambda=0.1$, the clipping parameter to $\epsilon=0.2$, and the KL penalty coefficient to $\beta=0.1$. These values remain fixed throughout training, without scheduling or adaptive adjustment.
For retrieval-augmented planning, we retrieve the top-$K$ manual sections with $K=5$, unless otherwise specified.

\paragraph{\uitars}
We fine-tune UI-TARS-1.5-7B using LoRA~\citep{DBLP:journals/corr/abs-2106-09685} with rank 16 and scaling parameter $\alpha=32$, applied to the query, value, and output projection layers.
We use the Adam optimizer with a learning rate of $1 \times 10^{-5}$.
Training is conducted over two rounds.
In the first round, we generate three rollouts per task and train for two epochs.
In the second round, we generate two rollouts per task and again train for two epochs.

\paragraph{\evocua}
We fine-tune EvoCUA-8B using LoRA with rank 8 and scaling parameter $\alpha=16$, applied to the query, value, and output projection layers.
We use the Adam optimizer with a learning rate of $1 \times 10^{-6}$.
We observed that EvoCUA tends to overfit more quickly than UI-TARS-1.5 under our training setup, and therefore conduct only a single training round for EvoCUA.
In this round, we generate five rollouts per task and train for two epochs.

\subsection{Computational Cost for \vscode}
\label{appendix:cost_vscode}
Unlike the other applications, the \vscode experiments were conducted on six NVIDIA RTX 4090 GPUs due to hardware availability.
Adapting \ours to \vscode incurred an API cost of \$91.25, with 73.43 hours required for episode generation and 34.65 hours for policy optimization, resulting in a total wall-clock time of 108.08 hours. 
Because the \vscode experiments were conducted under a different GPU configuration from the H100-based experiments, we report these computational costs separately rather than directly comparing their wall-clock time with those of the other applications.

\section{Slack Test Set Construction and Robustness}
\label{appendix:slack_eval}

This section describes the construction of the Slack test set
and evaluates the robustness of our results to task structure.

\paragraph{Slack test set construction}
To minimize potential bias toward the training data, we construct the Slack test set independently of the user manuals used for training.
We define 30 evaluation tasks across 15 functional themes, including messaging, channel management, and workspace administration.
For tasks requiring an initial setup, we use GPT-5.4 to construct corresponding setup tasks.
The resulting test set consists of 18 main-only tasks and 12 setup--main task pairs.

\begin{table}[t]
\centering
\small
\caption{Slack evaluation with and without setup tasks.}
\label{tab:slack_setup_free}

\resizebox{0.55\linewidth}{!}{%
\begin{tabular}{l|cc}
\toprule
& \multicolumn{2}{c}{\textbf{Gemini 3.1 Pro Preview}} \\
\cmidrule(lr){2-3}
\textbf{Method}
& \textbf{w/o Setup (18 tasks)}
& \textbf{All (30 tasks)} \\
\midrule
UI-TARS
& $27.78 \pm 6.80$
& $26.67 \pm 5.27$ \\

\rowcolor{blue!15}
MAGE
& $\mathbf{56.67 \pm 6.09}$
& $\mathbf{49.33 \pm 11.88}$ \\
\bottomrule
\end{tabular}%
}
\end{table}

\paragraph{Setup-free evaluation}
We further examine whether the performance gains depend on the setup--main task structure.
To this end, we evaluate the 18 tasks that require no preceding setup.
As shown in Table~\ref{tab:slack_setup_free}, \ours continues to outperform \uitars on this subset.
These results suggest that the improvements also extend to tasks without a separate setup stage.

\section{Analysis of Retrieval-Augmented Planning}
\label{appendix:Sensitivity}

Unless otherwise specified, all analyses in this section are conducted on Calc (OSWorld) using \ours with UI-TARS-1.5 as the backbone.
Results for the backbone model and \ours are reported as the mean over five independent runs, while all other variants are evaluated using a single run.
The ``Without RAP'' setting uses the adapted CUA without a plan during evaluation and corresponds to the PO-only setting in Table~\ref{tab:ablation}.
We analyze the sensitivity and robustness of Retrieval-Augmented Planning (RAP) by systematically varying its key components and configurations and examining their effects on task performance.

\subsection{Manual Quality Sensitivity}

\begin{table*}[!t]
\centering

\begin{minipage}[t]{0.48\textwidth}
\centering
\caption{Sensitivity to manual quality on Calc.
Old uses the 2010 manual instead of the 2021 manual, while Partial uses a subset of the 2021 manual.}
\label{tab:manual_quality}
\centering
\small
\setlength{\tabcolsep}{8pt}

\begin{tabular}{l|c}
\toprule
\textbf{Manual} & \textbf{SR (\%)} \\
\midrule
\textbf{Full (2021)} & \textbf{34.04} \\
Old (2010) & 25.53 \\
Partial (2021, 75\%) & 25.53 \\
Partial (2021, 50\%) & 21.28 \\
Partial (2021, 25\%) & 17.02 \\
Without RAP & 14.89 \\
\bottomrule
\end{tabular}

\end{minipage}
\hfill
\begin{minipage}[t]{0.48\textwidth}
\centering

\caption{Sensitivity of MAGE to retrieval quality on Calc under Top-5, lower-ranked, random, and without RAP.}
\label{tab:sen_retrieval}
\centering
\small
\setlength{\tabcolsep}{5pt}

\begin{tabular}{l|c}
\toprule
\textbf{Retrieval Method} & \textbf{SR (\%)} \\
\midrule
\textbf{Top-5 retrieval (Ours)} & \textbf{34.04} \\
Random 5 sections & 27.66 \\
Low-ranked sections (Ranks 6--10) & 14.89 \\
Without RAP & 14.89 \\
\bottomrule
\end{tabular}
\end{minipage}

\vspace{-0.8em}
\end{table*}

We evaluate \ours's robustness to documentation quality by introducing two forms of degradation: outdated and incomplete manuals.
We replace the LibreOffice 7.1 manual (2021) with the 3.3 version (2010) to test age,\footnote{The LibreOffice manuals used in this experiment are available online: \href{https://documentation.libreoffice.org/assets/Uploads/Documentation/en/CG7.1/CG71-CalcGuide.pdf}{LibreOffice 7.1 Calc Guide} and \href{https://wiki.documentfoundation.org/images/7/79/0105GS3-GettingStartedWithCalc.pdf}{LibreOffice 3.3 Calc Guide}.} and randomly retain 75\%, 50\%, and 25\% of the 2021 manual's sections to test completeness.
All evaluations are conducted on Calc.

As Table~\ref{tab:manual_quality} indicates, performance scales with manual quality. 
Using the older 2010 manual drops the success rate from 34.04\% to 25.53\%, and performance decreases as fewer sections of the 2021 manual are retained, reaching 17.02\% at 25\% retention.
Importantly, all documentation-backed setups outperform the baseline lacking retrieval-augmented planning (RAP) (14.89\%). This confirms that while \ours optimally leverages comprehensive, up-to-date text, it still extracts actionable procedural guidance from suboptimal sources.

\subsection{Retrieval Sensitivity}
We analyze the sensitivity of \ours to two retrieval factors: retrieval quality and the number of retrieved manual sections (Top-$K$). For retrieval quality, we compare the default Top-5 retrieval with lower-ranked sections (Ranks 6--10), five randomly selected sections, and a setting without retrieval-augmented planning (RAP). For Top-$K$, we vary the number of retrieved sections. All experiments are conducted on Calc under the same evaluation setting.

\paragraph{Retrieval quality} 
As shown in Table~\ref{tab:sen_retrieval}, performance drops to 27.66\% with randomly selected sections and further decreases to 14.89\% with lower-ranked sections (Ranks 6--10). 
This indicates that MAGE is sensitive to retrieval quality and that retrieving relevant manual sections has a substantial impact on its overall performance.

\begin{table*}[!t]
\centering

\begin{minipage}[t]{0.38\textwidth}
\centering
\captionof{table}{
Comparison of Top-$K$ retrieval
}
\label{tab:topk_sensitivity}
\small
\setlength{\tabcolsep}{10pt}

\begin{tabular}{l|c}
\toprule
\textbf{Top-$K$} & \textbf{SR (\%)} \\
\midrule
Top-1                  & 25.53 \\
Top-3                  & 27.66 \\
\textbf{Top-5 (Ours)}  & \textbf{34.04} \\
Top-10                 & 25.53 \\
Without RAP            & 14.89 \\
\bottomrule
\end{tabular}
\end{minipage}%
\hfill
\begin{minipage}[t]{0.58\textwidth}
\captionof{table}{Human validation of generated tasks in terms of usefulness, feasibility, and visual verifiability.}
\label{tab:human_tasks}
\centering
\small
\setlength{\tabcolsep}{4pt}

\resizebox{\linewidth}{!}{%
\begin{tabular}{c|ccc}
\toprule
\textbf{Program} & \textbf{Usefulness} & \textbf{Feasibility} & \textbf{Visual Verifiability} \\
\midrule
Calc    & 98.0\% (49/50) & 80.0\% (40/50) & 100.0\% (50/50) \\
Impress & 86.0\% (43/50) & 84.0\% (42/50) & 100.0\% (50/50) \\
Writer  & 96.0\% (48/50) & 88.0\% (44/50) & 100.0\% (50/50) \\
\midrule
\textbf{Overall}
& \textbf{93.3\% (140/150)}
& \textbf{84.0\% (126/150)}
& \textbf{100.0\% (150/150)} \\
\bottomrule
\end{tabular}%
}
\end{minipage}

\vspace{-0.8em}
\end{table*}

\paragraph{Effect of top-$K$} 
As shown in Table~\ref{tab:topk_sensitivity}, Top-5 retrieval achieves the highest observed success rate (34.04\%) among the evaluated settings. Increasing $K$ further to 10 reduces the success rate to 25.53\%.
This suggests that retrieving few sections may provide insufficient procedural information, whereas retrieving many sections can introduce less relevant content that interferes with planning. 
Based on this analysis, we use Top-5 retrieval as the default configuration throughout our experiments.

\section{Human Validation of \ours Components}
We conduct a human validation study to assess the reliability of the LLM-generated supervision used in \ours. Specifically, we examine three components: generated tasks, manual-guided plans, and trajectory evaluation.

\subsection{Human Validation of Generated Tasks}

We manually inspect 150 generated tasks, including 50 tasks each from Calc, Impress, and Writer. Each task is evaluated according to three binary criteria:

\begin{enumerate}
    \item \textbf{Usefulness:} whether the task corresponds to a meaningful operation that a real user may perform.
    \item \textbf{Feasibility:} whether the instruction is sufficiently specified and executable in the given environment.
    \item \textbf{Visual verifiability:} whether successful task completion can be determined from the resulting screenshots.
\end{enumerate}

As shown in Table~\ref{tab:human_tasks}, 93.3\% of the generated tasks are judged useful, and all tasks are visually verifiable. The feasibility rate is lower at 84.0\%, with the remaining cases primarily involving ambiguous instructions or missing prerequisites, such as required files or objects. These results show that the generation process produces largely meaningful and visually verifiable tasks, while also motivating the use of filtering before episode generation to remove infeasible instances.

\begin{table*}[!t]
\centering

\begin{minipage}[t]{0.53\textwidth}
\captionof{table}{Human evaluation of retrieval-augmented plans.}
\label{tab:human_val_plans}
\centering

\small
\setlength{\tabcolsep}{10pt}

\begin{tabular}{l|c}
\toprule
\textbf{Criterion} & \textbf{Pass Rate} \\
\midrule
Requirement coverage           & 82.98\% (39/47) \\
Procedural correctness         & 72.34\% (34/47) \\
Executability                  & 76.60\% (36/47) \\
Scope safety                   & 93.62\% (44/47) \\
\midrule
\textbf{Helpful for task execution} & \textbf{70.21\% (33/47)} \\
\bottomrule
\end{tabular}
\end{minipage}
\hfill
\begin{minipage}[t]{0.43\textwidth}
\captionof{table}{Agreement between GPT-5.4 and the official OSWorld evaluator.}
\label{tab:human_val_traj}
\centering

\small
\setlength{\tabcolsep}{10pt}

\begin{tabular}{c|c}
\toprule
\textbf{Program} & \textbf{Agreement} \\
\midrule
Calc        & 82.98\% (39/47) \\
Impress     & 73.33\% (33/45) \\
Writer      & 86.36\% (19/22) \\
\midrule
\textbf{Overall} & \textbf{79.82\% (91/114)} \\
\bottomrule
\end{tabular}
\end{minipage}

\vspace{-0.8em}
\end{table*}

\subsection{Human Validation of Generated Plans}

We further evaluate the quality of manual-guided plans on OSWorld Calc using four criteria: \textbf{requirement coverage} (covering key task requirements), \textbf{procedural correctness} (following a valid procedure), \textbf{executability} (being actionable in the environment), and \textbf{scope safety} (avoiding unnecessary actions). Minor correctable errors are tolerated, while missing essential steps or task-blocking errors are considered failures.

As shown in Table~\ref{tab:human_val_plans}, a plan is considered helpful if it satisfies requirement coverage and at least two of the remaining three criteria. Under this definition, 70.21\% of the generated plans are helpful, indicating that most plans capture the task objective without introducing substantial out-of-scope actions.

\subsection{External Validation of the VLM Evaluator}
\label{appendix:evaluator_validation}
To validate the VLM-based evaluator, we measured its agreement with OSWorld's rule-based evaluation functions. Table~\ref{tab:human_val_traj} shows an overall agreement of 79.82\% (91/114) across Calc, Impress, and Writer. 
Disagreements mainly occur in multi-condition tasks where modifications are not fully verifiable from final screenshots. 
These results indicate substantial agreement with the official evaluator, while highlighting limitations in visually ambiguous cases.

\section{Action Space of CUAs}
\label{sec:actionspace}
\begin{table*}[t]
\caption{Native action space of the EvoCUA agent.}
\label{tab:evocua_action_space}
\centering
\small
\setlength{\tabcolsep}{5pt}
\renewcommand{\arraystretch}{1.1}

\begin{tabularx}{\textwidth}{l l X l}
\toprule
\textbf{Category} & \textbf{Action Primitive} & \textbf{Description} & \textbf{Required Arguments} \\
\midrule

\multirow{4}{*}{\textbf{Keyboard}}
& \texttt{key}
& Presses and releases the specified key(s).
& \texttt{keys} (array) \\

& \texttt{key\_down}
& Presses and holds the specified key(s).
& \texttt{keys} (array) \\

& \texttt{key\_up}
& Releases the specified key(s).
& \texttt{keys} (array) \\

& \texttt{type}
& Types the specified text.
& \texttt{text} (string) \\
\midrule

\multirow{8}{*}{\textbf{Mouse}}
& \texttt{mouse\_move}
& Moves the cursor to the specified coordinates.
& \texttt{coordinate} $(x, y)$ \\

& \texttt{left\_click}
& Performs a left click.
& \texttt{coordinate} $(x, y)$ \\

& \texttt{right\_click}
& Performs a right click.
& \texttt{coordinate} $(x, y)$ \\

& \texttt{middle\_click}
& Performs a middle click.
& \texttt{coordinate} $(x, y)$ \\

& \texttt{double\_click}
& Performs a left double-click.
& \texttt{coordinate} $(x, y)$ \\

& \texttt{triple\_click}
& Performs a left triple-click.
& \texttt{coordinate} $(x, y)$ \\

& \texttt{left\_click\_drag}
& Clicks and drags to the specified coordinates.
& \texttt{coordinate} $(x, y)$ \\

& \texttt{scroll / hscroll}
& Performs vertical or horizontal scrolling.
& \texttt{pixels} (number) \\
\midrule

\multirow{2}{*}{\textbf{Control}}
& \texttt{wait}
& Pauses execution for the specified duration.
& \texttt{time} (number) \\

& \texttt{terminate}
& Terminates the task with the specified status.
& \texttt{status} (\texttt{"success"}$\mid$\texttt{"failure"}) \\
\bottomrule
\end{tabularx}

\vspace{-0.8em}
\end{table*}
\begin{table}
\caption{Action space of the UI-TARS-1.5~\citep{uitars} agent.}
\label{tab:action_space}
\centering
\small
\setlength{\tabcolsep}{6pt}

\begin{tabular}{ll}
\toprule
\textbf{Action} & \textbf{Definition} \\
\midrule
\texttt{click(x, y)} & Clicks at coordinates $(x, y)$. \\
\texttt{left\_double(x, y)} & Double-clicks at coordinates $(x, y)$. \\
\texttt{right\_single(x, y)} & Right-clicks at coordinates $(x, y)$. \\
\texttt{drag(x1, y1, x2, y2)} & Drags from $(x_1, y_1)$ to $(x_2, y_2)$. \\
\texttt{hotkey(key)} & Presses the specified keyboard shortcut. \\
\texttt{type(content)} & Types the specified text content. \\
\texttt{scroll(x, y, direction)} & Scrolls at $(x, y)$ in the specified \texttt{direction}$\in\{\texttt{up}, \texttt{down}\}$. \\
\texttt{wait()} & Waits for a brief moment. \\
\texttt{finished()} & Marks the task as complete. \\
\bottomrule
\end{tabular}

\vspace{0.05in}
\end{table}

\paragraph{\uitars}
The action space of \ours follows the action space of \uitars.
Detailed descriptions of the action space are provided in Table~\ref{tab:action_space}.

\paragraph{\evocua}
The action space of \ours with the EvoCUA backbone follows the action space defined by EvoCUA.
Detailed descriptions of the action space are provided in Table~\ref{tab:evocua_action_space}.

\section{Prompts}
\subsection{Manual-Derived Task Generation Prompt}
\label{app:prompt_task_generation}

We use the following prompt to generate manual-derived task samples from each manual section.
The same prompt template is used across applications, while the input program name, manual section text, and section images are replaced for each section.
Placeholders enclosed in braces denote instance-specific inputs.
\begin{tcolorbox}[title=Manual-Derived Task Generation Prompt, breakable]
\begin{lstlisting}
You are an expert at generating GUI-based practice tasks for AI agents.

You will be given:
1. Program name
2. One section of manual text, which may span multiple pages
3. One or more images from the manual section

Your task is to generate executable GUI practice tasks based strictly on the given manual section.

Each task must be output as an instruction pair with:
- setup instruction, if necessary
- main instruction

Each instruction must contain:
- task: a short sentence describing the goal
- steps: a list of concrete GUI actions

Rules:

1. Generate up to two task samples, or none if the section is irrelevant.

2. The task must produce a visible result on the screen.
   Do not generate tasks such as:
   - opening menus only
   - checking whether a feature exists
   - saving a file
   - exploring the UI
   The result must be visually observable, such as formatted text, a created chart, a changed slide object, or hidden columns.

3. The starting screen is always a freshly opened blank/default screen.
   No files, tables, documents, slides, conversations, or other pre-existing content are assumed.

4. The GUI agent must not be asked to make subjective or free choices.
   Do not generate steps such as:
   - choose any color
   - select a preferred style
   - pick an option you like
   - apply a random format
   All actions must be deterministic and explicitly specified.

5. Output format:

[
  {
    "setup_task": "...",
    "setup_steps": [...],
    "main_task": "...",
    "main_steps": [...],
    "id": 0
  }
]

6. Setup instruction:
- Setup is only for creating or preparing data or objects required by the main task.
- If the main task needs any pre-existing value, text, table, slide object, or conversation content, it must be created in setup_steps.
- Main steps must never include data creation actions such as typing initial values, creating tables, or preparing sample objects.
- If the main task can be done from a blank screen without preparation, leave setup_task as an empty string and setup_steps as an empty list.

7. Main instruction:
- Main steps must only apply the manual feature to the prepared data or objects.
- Main steps must not include:
  - entering initial numbers or text
  - creating tables
  - preparing sample data
  - creating initial objects
- Main task must be a concrete exercise reconstructed from the manual section.

8. If sample data or objects are required in setup:
   - Do not rely only on images.
   - Describe exactly how to create the required data or objects in text steps.

9. Steps must:
   - be atomic GUI actions, such as click, type, select, drag, or press keys
   - be ordered
   - not include explanations
   - not mention the manual or images

10. Do not invent features not mentioned in the given manual section.

11. Section-based multimodal input handling:
   - The text is section-level, not single-page.
   - Multiple images may be provided for the same section.
   - Use all provided images as supporting context when relevant.
   - If text and images conflict, prioritize the manual text.
   - Do not assume hidden UI details not visible in the text or images.

12. Output must be only valid JSON.
   - No markdown
   - No explanation
   - No extra text

Input:
Program name:
{program_name}

Manual section text:
{manual_section_text}

Manual section images:
{manual_section_images}
\end{lstlisting}
\end{tcolorbox}

\subsection{Manual-Guided Plan Generation Prompt}
\label{sec:plannerprompt}
This prompt is used to generate a task-specific plan during Manual-Guided Planning.
It takes the task instruction, the initial screenshot, and retrieved manual sections as input, and outputs a concise step-by-step plan that guides the CUA during action generation.
\label{app:prompt_plan_generation}
\begin{tcolorbox}[title=Manual-Guided Plan Generation Prompt, breakable]
\begin{lstlisting}
You are a planner for a GUI agent operating on real screenshots.

You will be given:
1) A task instruction
2) The initial screenshot of the environment
3) Potentially relevant manual snippets retrieved for this task

Your goal is to generate a concise, step-by-step plan that the agent can follow to complete the task.

Requirements:
- Focus only on actions that are directly useful for completing the task.
- Each step must correspond to a realistic GUI interaction, such as click, type, scroll, or select.
- Ground each step in visible UI elements from the initial screenshot, such as button labels, menus, input fields, icons, or relative positions.
- Prefer explicit interaction targets, such as clicking the Insert menu at the top toolbar, instead of vague descriptions.
- If multiple paths are possible, choose the most straightforward and reliable one.
- Avoid assumptions about hidden UI or unavailable elements.
- If the UI element is not clearly visible, describe how to locate it, such as via menu navigation or scrolling.
- Do not include unnecessary explanations, reasoning, or background knowledge.
- Do not repeat the task instruction.

Robustness:
- If an action might fail, suggest a simple alternative interaction, such as using a button instead of typing.
- Ensure the plan is executable even with minor UI variations.

Output format:
- Output only one section titled "Relevant plan".
- Use short, ordered, numbered steps.
- Each step should describe exactly one action.

Input:
Task instruction:
{task_instruction}

Initial screenshot:
{initial_screenshot}

Retrieved manual sections:
{retrieved_manual_sections}
\end{lstlisting}
\end{tcolorbox}

\subsection{Trajectory Evaluation Prompt}
\label{app:prompt_evaluation}
The evaluator outputs both a binary success label and a continuous progress score.
The progress score is used only for consistency checking and is not used as a training reward.
For policy optimization, we assign rewards based on the binary success label.
\begin{tcolorbox}[title=Trajectory Evaluation Prompt, breakable]
\begin{lstlisting}
You are an evaluation agent.

You will be provided with:
1. The Task objective
2. The last 5 images of the task trajectory.

Your tasks consist of two parts:

PART 1: Task Success Evaluation
- Determine whether the Task objective has been successfully completed.
- Evaluate this by referring to:
  - the visual evidence in the last 5 step images
- Do not rely only on the final image; consider transitions and changes across all five images.
- Carefully analyze visual information and logical progression step-by-step.

PART 2: Task Progress Evaluation
- Evaluate how much the Task has progressed toward completion.
- Use:
  - the visual evidence across the last 5 step images
- Do NOT rely only on the final image.
- Compare observed behavior strictly against the Task requirements.
- The score should reflect how much of the Task has been completed so far, not just whether it succeeded or failed.

Scoring Guidelines:
- Assign a score between 0.0 and 1.0 in increments of 0.01.
- 0.00 = no meaningful progress toward the Task goal
- 1.00 = Task fully completed
- Intermediate values should represent proportional progress.
- Consider:
  - how many required steps have been completed
  - how correct the completed steps are
  - whether errors block further progress

Output Rules:
- Output MUST be valid JSON.
- Do NOT include any text outside the JSON object.
- Always output the score as a JSON number, not a string.

Output format:
{
  "result_thought": "Explain, step-by-step, how you analyzed the last 5 images and why you concluded that the Task succeeded or failed.",
  "result": "SUCCESS" or "FAIL",
  "score_thought": "Explain, step-by-step, how you analyzed the last 5 images and why you assigned this progress score.",
  "score": 0.00
}
\end{lstlisting}
\end{tcolorbox}

\lstset{
  basicstyle=\ttfamily\small,
  breaklines=true,
  columns=fullflexible,
  keepspaces=true,
  showstringspaces=false
}
\end{document}